\documentclass[11pt,a4paper]{article}
\usepackage[hyperref]{acl2018}
\usepackage{times}
\usepackage{latexsym}

\usepackage{url}

\usepackage{multirow}
\usepackage{amsthm}
\usepackage{amssymb}
\usepackage{amsfonts}
\usepackage{color}
\usepackage{MnSymbol}
\usepackage{makecell}
\usepackage{arydshln}
\usepackage{graphicx}
\usepackage{microtype}
\usepackage[shortlabels]{enumitem}

\usepackage{subcaption}

\usepackage{caption} 
\usepackage{booktabs}

\usepackage{algorithm}
\usepackage[noend]{algpseudocode}

\algdef{SE}[SUBALG]{Indent}{EndIndent}{}{\algorithmicend\ }%
\algtext*{Indent}
\algtext*{EndIndent}

\aclfinalcopy 
\def\aclpaperid{***} 
 
\title{Seq2Seq2Sentiment: \\ Multimodal Sequence to Sequence Models for Sentiment Analysis}

\author{Hai Pham$^1$*, Thomas Manzini$^1$*, Paul Pu Liang$^2$, Barnab\'{a}s Pocz\'{o}s$^2$ \\
\{$^1$Language Technologies Institute, $^2$Machine Learning Department\}, CMU, USA \\ {\tt \{htpham,tmanzini,pliang,bapoczos\}@cs.cmu.edu} \\}
\date{}

\begin{document}
\maketitle

\begin{abstract}
Multimodal machine learning is a core research area spanning the language, visual and acoustic modalities. The central challenge in multimodal learning involves learning representations that can process and relate information from multiple modalities. In this paper, we propose two methods for unsupervised learning of joint multimodal representations using sequence to sequence (Seq2Seq) methods: a \textit{Seq2Seq Modality Translation Model} and a \textit{Hierarchical Seq2Seq Modality Translation Model}. We also explore multiple different variations on the multimodal inputs and outputs of these seq2seq models. Our experiments on multimodal sentiment analysis using the CMU-MOSI dataset indicate that our methods learn informative multimodal representations that outperform the baselines and achieve improved performance on multimodal sentiment analysis, specifically in the Bimodal case where our model is able to improve F1 Score by twelve points. We also discuss future directions for multimodal Seq2Seq methods.

\end{abstract}

\section{Introduction}

Sentiment analysis, which involves identifying a speaker's sentiment, is an open research problem. 
In this field, the majority of work done focused on unimodal methodologies - 
primarily textual analysis - where investigating was limited to identifying usage of words in positive and negative scenarios. 
However, unimodal textual sentiment analysis through usage of words, phrases, and their interdependencies were found to be insufficient for extracting affective content from textual opinions \citep{rosas2013multimodal}.
\footnote{*These authors contributed equally.}
As a result, there has been a recent push towards using statistical methods to extract additional behavioral cues not present in the language modality from the video and audio modalities. This research field is known as multimodal sentiment analysis and it extends the conventional text-based definition of sentiment analysis to a multimodal setup where different modalities contribute to modeling the sentiment of the speaker. For example, \citep{kaushik2013sentiment} explores modalities such as audio, while \citep{wollmer2013youtube} explores a multimodal approach to predicting sentiment. This push has been further bolstered by the advent of multimodal social media platforms, such as YouTube, Facebook, and VideoLectures which are used to express personal opinions on a worldwide scale. 
As a result, several multimodal datasets, such as CMU-MOSI \citep{zadeh2016multimodal} and later CMU-MOSEI \citep{cmumoseiacl2018}, ICT-MMMO \citep{wollmer2013youtube}  and YouTube \citep{morency2011towards}, take advantage of the abundance of multimodal data on the Internet. 
At the same time, neural network based multimodal models have been proposed that are highly effective at learning multimodal representations for multimodal sentiment analysis \citep{chen2017msa,poria2017context,zadeh2018memory,zadeh2018multi}.

Recent progress has been limited to supervised learning using labeled data, and does not take advantage of the abundant unlabeled data on the Internet. 
To address this gap, our work is primarily one of unsupervised representation learning. 
We attempt to learn a multimodal representation of our data in a structured paradigm and explore whether a joint multimodal representation trained via unsupervised learning can improve the performance for multimodal sentiment analysis. 
While representation learning has been an area of rapid research in the past years, there has been limited work that explores multimodal setting. 
To this end, we propose two methods: a \textit{Seq2Seq Modality Translation Model} and a \textit{Hierarchical Seq2Seq Modality Translation Model} for unsupervised learning of multimodal representations. Our results show that using multimodal representations learned from our Seq2Seq modality translation method outperforms the baselines and achieves improved performance on multimodal sentiment analysis.

\section{Related Work}

In the past, approaches to text-based emotion and sentiment recognition rely mainly on rule-based techniques, bag of words (BoW) modeling or SNoW architecture \citep{Chaumartin_2007} using a large sentiment or emotion lexicon \citep{mishne2005experiments}, or statistical approaches that assume the availability of a large dataset annotated with polarity or emotion labels.



Multimodal sentiment analysis has gained a lot of research interests over the last few years \citep{mm_survey}. Probably the most challenging task in multimodal sentiment analysis is to find a joint representation of multiple modalities. This problem is has been approached in a number of ways. Earlier works such as \citep{ngiam2011multimodal,lazaridou2015combining, kiros2014unifying} have pushed some progress towards this direction. 

Recently, more advanced neural network models were proposed to learn multimodal representations. The Multi-View LSTM (MV-LSTM) \citep{Rajagopalan2016} was suggested to exploit fusion and temporal relationships. MV-LSTM partitions memory cells and gates into multiple regions corresponding to different views. Tensor Fusion Network \citep{tensoremnlp17} presented an efficient method based on Cartesian-product to take into consideration intramodal and intermodal relations between video, audio and text of the reviews to create a novel feature representation for each utterance. The Gated Multimodal Embedding model \cite{chen2017msa} created an algorithm using reinforcement learning to train an on-off switch that decided what values the video and audio components would have. Noisy modalities are turned off and clean modalities are allowed to pass through. \cite{zadeh2018memory} utilizes external multimodal memory mechanisms to store multimodal information and create multimodal representations through time. \cite{zadeh2018multi} proposed using multiple attention coefficient assignments to represent multiple cross-modal interactions. However, all these methods discussed so far are purely supervised approaches to multimodal sentiment analysis and do not leverage the power of unsupervised data and generative approaches towards learning multimodal representations.

Besides supervised approaches, generative methods based on generative adversarial networks (GAN) \citep{gan} have attracted significant interest in learning joint distribution between two or more modalities \citep{bigan,triplegan,trianglegan}. Another method to deal with multimodal problems is to view them as conditional problems which learn to map a modality to the other \citep{conditionalgan,conditionalvae,variationalmultimodal}. Our work can be viewed as an extension of the conditional approach, as both utilize unsupervised learning. However, our work differs from those in that it takes into account the sequential dependency within each modality. 


Finally, attention based layers have also proved themselves to be effective tools to boost performance of neural network models, such as in neural machine translation \citep{2017opennmt,bahdanau2014neural,luong2015effective}, speech recognition \citep{coldfusion} and in image captioning \citep{xu2015show}. Our work also employs this mechanism in an attempt to better handle long-term dependencies of variable-length sequences. 

\section{Problem Formulation}

Given a dataset with data $X = (X^{text}, X^{audio}, X^{video})$ where $X^{text}$, $X^{audio}$, $X^{video}$ stand for text, audio and video modality inputs, respectively. Typically a dataset is indexed by videos. This means that if we have $n$ videos, then $X = (X_1, X_2, ..., X_n)$ where $X_i = ({X_i}^{text}, {X_i}^{audio}, {X_i}^{video}), \, 1 \leq i \leq n$. The corresponding labels for these $n$ videos are $Y=(Y_1, Y_2, ..., Y_n), \, Y_i \in \mathbb{R}$. 

To simplify the problem, we align the input based on words. Typically, researchers often segment each video into a smaller set in which each segmented video will last a couple of seconds, instead of minutes as done in ~\cite{chen2017msa}. After such alignment and segmentation, we have the equal-length inputs of each modality per video. For example, at the $i^{th}$ video, we have ${X_i}^{text} = ({w_i}^{(1)}, {w_i}^{(2)}, ..., {w_i}^{(T_i)})$ where ${w_i}^{(t)}$ stands for the $t^{th}$ word and $T_i$ is the length of the $i^{th}$ video's text input, \textit{a.k.a} time steps. Note that different videos will have different time steps. Similarly for this video, we have a sequence of audio input ${X_i}^{audio} = ({a_i}^{(1)}, {a_i}^{(2)}, ..., {a_i}^{(T_i)})$ and video input ${X_i}^{video} = ({v_i}^{(1)}, {v_i}^{(2)}, ..., {v_i}^{(T_i)})$. 

In this work we are tackling the input learning problem where we want to learn the embedding representation for all text, audio, and video modalities: $\widetilde{X_i} = f(X_i) = f(({X_i}^{text}, {X_i}^{audio}, {X_i}^{video})) $.
In our baseline model, the function $f$ is simply the concatenation at time step level: $\widetilde{x_i}^t = [ {w_i}^{t}; {a_i}^{t} ; {v_i}^{t} ] $

In our proposed method, we learn $\widetilde{X_i}$ by using a Seq2Seq model. We do not calculate each embedding representation for each time step, but for the whole sequence. Formally, $\widetilde{X_i} = f(X_i) = Seq2Seq\_Encoder(X_i)$ where $Seq2Seq\_Encoder$ is the encoder part of our Seq2Seq model. 

Now, we have the transformed inputs $\widetilde{X} = (\widetilde{X_1}, \widetilde{X_2}, ..., \widetilde{X_n})$ and outputs $Y=(Y_1, Y_2, ..., Y_n)$ for $n$ videos, where $\widetilde{X_i} = (\widetilde{x_i}^1, \widetilde{x_i}^2, ..., \widetilde{x_i}^{T_i})$. For simplicity, in the next formula, we omit the index of video segment $i$, and so the input becomes $\widetilde{X} = (\widetilde{x}^1, \widetilde{x}^2, ..., \widetilde{x}^{T})$, and the labels become ${Y} = ({y}^1, {y}^2, ..., {y}^{T})$.

We will be using a Recurrent Neural Network (RNN) such as LSTM \citep{lstm} or GRU \citep{gru} to model this sequence. In detail, this RNN has a stack of $K$ hidden layers $h = ({h^1, h^2, ..., h^K})$, each contains $D$ 
hidden neurons: ${h}^k=({h_1^k, h_2^k, ..., h_D^k}), \, k\in[1,K]$. We denote ${W}$ and ${b}$ to be weight and bias, then for the first layer which contacts directly with input: 
  \begin{equation}
      {h^1}_t = H(W_{x{h^1}}{\widetilde{x_t}} + 
                  W_{{h^1}{h^1}}{h^1}_{t-1} + 	{b}_{h^1})
  \end{equation}
where $H$ is the RNN cell function. For example of LSTM, it contains \textit{input, forget, output} and \textit{cell state}. At hidden layer $k \in [2, K]$: 
  \begin{equation}
      {h^k}_t = H(W_{{h^{t-1}}{h^t}}{h^{k-1}_t} + 
                  W_{{h^k}{h^k}}{h^k}_{t-1} + {b}_{h^k})
  \end{equation}
Optionally, we apply a soft attention mechanism \textit{on top} of the last hidden layer ${h^K}$, with shared weight $W_{\alpha}$ over $T$ time steps, then we can obtain the attention output $\alpha$:
  \begin{equation}
  \alpha = softmax \begin{pmatrix}
                      \begin{bmatrix} 
                         W_{\alpha}h_1^K \\
                         W_{\alpha}h_2^K \\
                         ... \\
                         W_{\alpha}h_T^K \\
                      \end{bmatrix}
                   \end{pmatrix}\
  \end{equation}
The last hidden layer's output now becomes: 
  \begin{equation}
      A = [h_1^K, h_2^K, ..., h_T^K] \alpha = H^K \alpha
  \end{equation}
  And the last output layer with regression score is: 
  \begin{equation}
  \begin{aligned}
      & \widetilde{y_t} = {W_{Ay}}{A} + b_y
      \\
  \end{aligned}
  \end{equation}

Finally, we calculate the loss with respect to the labels. As in~\cite{chen2017msa}, we choose Mean Absolute Error (MAE) as our loss and later train with stochastic gradient descent: 
  \begin{equation}
      \mathbb{L}_{MAE}(\widetilde{{Y}}, {{Y}}) =  
      \mathbb{E}[|\widetilde{Y} - Y|]
  \end{equation}
\begin{figure}[tbp]
      \centering
      \includegraphics[width=0.4\textwidth]{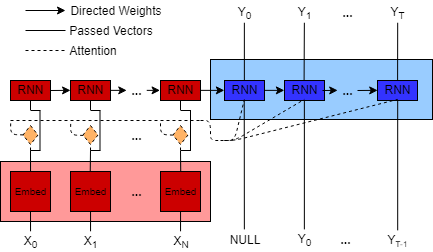}
\caption{\small{Seq2Seq Modality Translation Model
with input $(X_1, ..., X_N)$ and output is $(Y_1, ..., Y_T)$. 
Seq2Seq makes use of the whole input sequence in the decoding phase for every token $Y_i$. 
If attention model (yellow color) is used, for each $Y_i$, it learns a separate weight vector \textit{w.r.t} each token of input $X$ to see which token should the decoder ``attend'' more. 
      }}
      \label{fig:pretrain}
\end{figure}

  \begin{figure*}[!ht]
      \centering
      \includegraphics[width=0.7\textwidth]{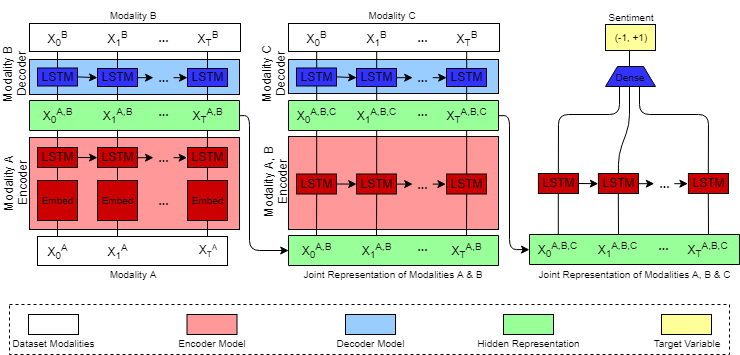}
      \caption{\small{Hierarchical Seq2Seq Modality Translation Model: first we train with 2 modalities, then we add one more on the second phase, from which the results will be fed into RNN for sentiment prediction. 
      The green boxes denote the joint representation learned by Seq2Seq models: the joint representation of modalities A and B will be fed into another Seq2Seq model which in turn learns the joint representation of AB and another modality C. Finally the joint representation of ABC will be fed into a RNN to predict sentiment.}}
      \label{fig:Seq2Seq}
  \end{figure*}


%
\section{Proposed Approach}\label{sec:PROPAPR}

In this section we describe the different approaches that we plan to take to improve affect recognition through learning multimodal representations.

\subsection{Seq2Seq Modality Translation Model}



The \textit{Seq2Seq Modality Translation Model} aims to learn multimodal representations that can be used for discriminative tasks. While Seq2Seq models have been predominantly used for machine translation \citep{bahdanau2014neural,luong2015effective}, we extend its usage to the realm of multimodal machine learning where we use it to translate one modality to another, or translate a joint representation to another single or joint representation. To do so, we propose a Seq2Seq modality translation model with attention mechanism, as shown in Figure \ref{fig:pretrain}. Modality $X$ is translated into modality $Y$. Our hypothesis is that the intermediate representation of this model, i.e. the output of Seq2Seq's encoder, or the input of its decoder, is close to the joint representation $(X,Y)$ of the two modalities involved. As a result, this representation can be used for tasks that involve learning joint representation across multiple modalities. The detail is in Algorithm \ref{alg:bimodal}. 

\begin{algorithm}
\small{
\caption{\textbf{Seq2Seq Modality Translation} \\ $X,Y,S$ are 2 modalities and sentiment sequences}
\label{alg:bimodal}
\begin{algorithmic}[1]
	\State \textbf{Phase 1: Train Seq2Seq}
    \Indent
		\State $\mathcal{E}_{XY} \gets Seq2Seq\_RNN\_Encode(X)$	
    	\State $\widetilde{Y} \gets Seq2Seq\_RNN\_Decode(\mathcal{E}_{XY})$ 
        \State $loss = cross\_entropy(\widetilde{Y}, Y)$
        \State $\text{Backprop to update params}$ 
	\EndIndent

\item[]

	\State \textbf{Phase 2: Sentiment Regression}
    \Indent
    	\State $\mathcal{E}_{XY} \gets Seq2Seq\_RNN\_Encode(X)$ 
			\Comment trained encoder in Seq2Seq model
        \State $R = RNN(\mathcal{E}_{XY})$
        \State $score \gets Regression(R)$
        \State $loss \gets MAE(score, S)$
        \State $\text{Backprop to update params}$
     \EndIndent
\end{algorithmic}
}
\end{algorithm}

Formally, the Seq2Seq Modality Translation Model consists of 2 separate steps: encoding and decoding, each phase typically consists of a single RNN or a stack of them. This model accepts variable-length inputs of $X$ and $Y$, and the network should be trained to maximize the translational condition probability $p(Y|X)$. For encoding, it encodes the whole input sequence X into an embedded representation. The hidden state output of each time step is based on the previous hidden state along with the input sequence (refer to Figure \ref{fig:pretrain}): 
  \begin{equation}
      h_n = RNN(h_{n-1}, X_n)	
  \end{equation}
The encoder's output is the final hidden state's output of the encoding RNN: 
  \begin{equation}
      \mathcal{E} = h_N = RNN(h_{N-1}, X_N)	
  \end{equation}
where N is the length of the input sequence X. The decoder tries to decode each token $Y_i$ at a time based on $\mathcal{E}$ and all previous decoded tokens, which is formulated as: 
  \begin{equation}
      p(Y) = \prod_{i=1}^{T} p(Y_i|\mathcal{E}, Y_1, ..., Y_{i-1})
  \end{equation}
The Seq2Seq training target is to find the best translation sequence which is as close to the ground truth Y as possible, or formally:
  \begin{equation}
      \widehat{Y} = \operatorname*{arg\,max}_{Y} p(Y|X) 
  \end{equation}
And while there are some other search algorithms such as random sampling or greedy search to decode each token \cite{neubig2017neural}, we use the traditional beam search approach \cite{Seq2Seq}. 
\subsection{Hierarchical Seq2Seq Modality Translation Model}

The Seq2Seq Modality Translation Model only learns joint representation between 2 modalities $X$ and $Y$. While this might be a strong starting point, we believe an approach that captures the joint interactions between all different modalities $X,Y,Z$ is more effective in modeling the full distribution of the multimodal data and therefore more useful for regression or classification. In response, we propose the \textit{Hierarchical Seq2Seq Modality Translation Model} that learns a joint multimodal representation. Once the Seq2Seq Modality Translation Model is trained for 2 modalities $X$ and $Y$, we obtain the intermediate representation $\mathcal{E}_{XY}$ which is the joint representation of $(X,Y)$. $\mathcal{E}_{XY}$ is in turn treated as input sequence for the next Seq2Seq Modality Translation Model to decode the third modality $Z$. The final multimodal representation $\mathcal{E}_{XYZ}$ represents the joint representation of $(X,Y,Z)$. The Hierarchical Seq2Seq Modality Translation Model is described as in Algorithm \ref{alg:trimodal}. 
\begin{algorithm}
\small{
\caption{\textbf{Hierarchical Seq2Seq Modality Translation}: $X,Y,Z,S$ are 3 modalities and sentiment sequences}
\label{alg:trimodal}
\begin{algorithmic}[1]
	\State \textbf{Phase 1: Train Seq2Seq for 2 modalities}
    \Indent
		\State $\mathcal{E}_{XY} \gets Seq2Seq\_RNN\_Encode(X)$	
    	\State $\widetilde{Y} \gets Seq2Seq\_RNN\_Decode(\mathcal{E}_{XY})$ 
        \State $loss = cross\_entropy(\widetilde{Y}, Y)$
        \State $\text{Backpropagate to update parameters}$ 
	\EndIndent

\item[]

\State \textbf{Phase 2: Train Seq2Seq for 3 modalities}
    \Indent
		\State $\mathcal{E}_{XYZ} \gets Seq2Seq\_RNN\_Encode(\mathcal{E}_{XY})$ 
    	\State $\widetilde{Z} \gets Seq2Seq\_RNN\_Decode(\mathcal{E}_{XYZ})$ 
        \State $loss = cross\_entropy(\widetilde{Z}, Z)$
        \State $\text{Backpropagate to update parameters}$ 
	\EndIndent

\item[]

\State \textbf{Phase 3: Sentiment Regression}
    \Indent
    	\State $\mathcal{E}_{XYZ} \gets Seq2Seq\_RNN\_Encode(\mathcal{E}_{XY})$ 
        \State $R = RNN(\mathcal{E}_{XYZ})$
        \State $score \gets Regression(R)$
        \State $loss \gets MAE(score, S)$
        \State $\text{Backpropagate to update parameters}$
	\EndIndent
\end{algorithmic}
}
\end{algorithm}

This strategy is also illustrated in Figure \ref{fig:Seq2Seq}. The output of the second Seq2Seq model is the input of the last RNN model where we will train to predict regression sentiment scores. This last Seq2Seq model will be trained using MAE loss function and it perform subsequent regression process. 
\section{Experimental Setup}\label{sec:Expset}
We explored the applications of this model to the CMU-MOSI dataset \citep{zadeh2016multimodal}. We implemented a baseline LSTM model based off the work done in \citep{chen2017msa}. Our implementation uses 66.67\% of the data for training from which we take a 15.15\% held-out set for validation, and the remaining 33.33\% is used for testing.
Finally, we evaluated our proposed model against the baseline results generated by the implementation of \citep{chen2017msa}. Here we compared our results against the various multimodal configurations evaluating our performance using precision, recall, and F1 scores.
\subsection{Dataset and Input Modalities}\label{sec:MOSI}
The dataset that we use to explore applications of our model is the CMU Multimodal Opinion-level Sentiment Intensity dataset (CMU-MOSI). The dataset contains video, audio, and transcriptions of 89 different speakers in 93 different videos divided into 2199 separate opinion sentiments. Each video has an associated sentiment label in the range from -3 to 3. The low end of the spectrum (-3) indicates strongly negative sentiment, where as the high end of the spectrum indicates strongly positive sentiment (+3), and ratings of 0 indicate neutral sentiment. The CMU-MOSI dataset is currently subject to much research \citep{poria2017context,chen2017msa,zadeh2018memory,zadeh2018multi} and the current state of the art is achieved by \citep{poria2017context} with an F1 score of 80.3 using a context aware model across entire videos. The state of the art using only individual segments is achieved by \citep{zadeh2018memory} with an F1 score of 77.3.

With respect to raw features that are being given as inputs to our model, we perform feature extraction in the same manner as described in \citep{chen2017msa}. In the text domain, pretrained 300 dimensional GLoVe embeddings \citep{pennington2014glove} were used to represent the textual tokens. In the audio domain, low level acoustic features including 12 Mel-frequency cepstral coefficients (MFCCs), pitch tracking and voiced/unvoiced segmenting features \citep{drugman2011joint}, glottal source parameters \citep{childers1991vocal,drugman2012detection,alku1992glottal,alku1997parabolic,alku2002normalized}, peak slope parameters and maxima dispersion quotients \citep{kane2013wavelet} were extracted automatically using COVAREP \cite{degottex2014covarep}. Finally, in the video domain, Facet \cite{emotient} is used to extract per-frame basic and advanced emotions and facial action units as indicators of facial muscle movement \citep{ekman1992argument,ekman1980facial}.

In situations where the same time alignment between different modalities are required, we choose the granularity of the input to be at the level of words. The words are aligned with audio using P2FA \cite{P2FA} to get their exact utterance times. The visual and acoustic modalities are aligned to words using these utterance times.

\begin{table*}[!ht]
\small{
\begin{center}
\begin{tabular}{*8c}
\toprule
\multirow{2}{*}{Method} & \multirow{2}{*}{Feature} & \multicolumn{3}{c}{BINARY ($-1$, +1)} & \multicolumn{3}{c}{7-CLASS ($-3$, ..., +3)} \\
{} & {} & \multicolumn{1}{l}{Prec} & \multicolumn{1}{l}{Recall} & \multicolumn{1}{l}{F1} & \multicolumn{1}{l}{Prec} & \multicolumn{1}{l}{Recall} & \multicolumn{1}{l}{F1} \\
\midrule
\multirow{3}{*}{UniModal-Baseline} & \multicolumn{1}{l}{Text (T)} & \multicolumn{1}{l}{\textbf{0.77}} & \multicolumn{1}{l}{\textbf{0.76}} & \multicolumn{1}{l}{\textbf{0.76}} & \multicolumn{1}{l}{\textbf{0.32}} & \multicolumn{1}{l}{\textbf{0.35}} & \multicolumn{1}{l}{\textbf{0.33}} \\
& \multicolumn{1}{l}{Audio (A)} & \multicolumn{1}{l}{0.56} & \multicolumn{1}{l}{0.56} & \multicolumn{1}{l}{0.56} & \multicolumn{1}{l}{0.12} & \multicolumn{1}{l}{0.19} & \multicolumn{1}{l}{0.14} \\
& \multicolumn{1}{l}{Video (V)} & \multicolumn{1}{l}{0.57} & \multicolumn{1}{l}{0.47} & \multicolumn{1}{l}{0.48} & \multicolumn{1}{l}{0.12} & \multicolumn{1}{l}{0.19} & \multicolumn{1}{l}{0.12} \\

\bottomrule
\end{tabular}
\end{center}
\caption{Unimodal baseline results with 3 metrics: Precision, Recall and F-Score (F1)
}\label{tbl:unimodal_baselines}
}
\end{table*}
\subsection{Baselines}
We use a LSTM model implemented in 3 different ways (one for each different grouping of the modalities). 
First in the unimodal domain, we run sentiment regression based solely on one modality, second in the bimodal domain we change the input to the concatenation of any pair of modality, and finally in the trimodal domain we concatenate all three modalities. 
This baseline not only serves to act as a benchmark for comparing our results but also acts as a starting point for our code development. As such, any improvements in our metrics are strictly as a result of the representations that we have learned and not structural changes in our model.

\subsection{Multimodal Model Variations} \label{subsec:variations}
Throughout our experimentation, we apply the algorithms in Section \ref{sec:PROPAPR} with several intuitive variations of how to translate modalities. 
Below are all approaches that we try to 
maximize our chances of learning a strong representation. 

For bimodal, we translate one modality into another one. For example, A $\rightarrow$ V stands for translating from Audio to Video, and take the embedding state, which we refer to as \texttt{embed}(A+V), to predict sentiment. Here we employ the Seq2Seq Modality Translation Model mentioned in Algorithm \ref{alg:bimodal}.

For trimodal, there are a lot more variations as follows. First, since we have 3 different modality and Seq2Seq is only capable of translating one modality to another, we use the Hierarchical Seq2Seq Modality Translation Model which is mentioned in Algorithm \ref{alg:trimodal}, e.g. we translate from \texttt{T} to \texttt{A} to have the joint representation \texttt{embed}(T+A), and then continue the translation from \texttt{embed}(T+A) to the rest modality which is \texttt{V}, which in turn yields the joint representation \texttt{embed}(T+A+V) to make sentiment prediction. 

Second, we reuse the previous Seq2Seq Modality Translation Model to translate a concatenation of 2 modality to the rest, e.g. \texttt{concat}(T+V) to A, and vice versa, e.g. translating from A back to \texttt{concat}(T+V).

Finally, we still use the Seq2Seq Modality Translation Model to translate from a concatenation of 2 modality to another concatenation of other 2. With this setting, at least one modality is repeated, and base on many previous works and our experience, we tend to favor text modality (T) over the other two and make it repeated. 

\section{Results} \label{sec:Results}

\begin{table*}[!ht]
\small{
\begin{center}
\begin{tabular}{*8c}
\toprule
\multirow{2}{*}{Method} & \multirow{2}{*}{Feature} & \multicolumn{3}{c}{BINARY ($-1$, +1)} & \multicolumn{3}{c}{7-CLASS ($-3$, ..., +3)} \\
{} & {} & \multicolumn{1}{l}{Prec} & \multicolumn{1}{l}{Recall} & \multicolumn{1}{l}{F1} & \multicolumn{1}{l}{Prec} & \multicolumn{1}{l}{Recall} & \multicolumn{1}{l}{F1} \\

\midrule
\multirow{3}{*}{BiModal-Baseline} 
& \multicolumn{1}{l}{\texttt{concat}(T + V)} & \multicolumn{1}{l}{\textbf{0.78}} & \multicolumn{1}{l}{\textbf{0.67}} & \multicolumn{1}{l}{{0.55}} & \multicolumn{1}{l}{{0.01}} & \multicolumn{1}{l}{{0.16}} & \multicolumn{1}{l}{{0.05}} \\
& \multicolumn{1}{l}{\texttt{concat}(T + A)} & \multicolumn{1}{l}{0.44} & \multicolumn{1}{l}{0.66} & \multicolumn{1}{l}{0.53} & \multicolumn{1}{l}{0.02} & \multicolumn{1}{l}{0.15} & \multicolumn{1}{l}{0.04} \\
& \multicolumn{1}{l}{\texttt{concat}(A + V)} & \multicolumn{1}{l}{0.55} & \multicolumn{1}{l}{0.47} & \multicolumn{1}{l}{0.48} & \multicolumn{1}{l}{0.13} & \multicolumn{1}{l}{0.16} & \multicolumn{1}{l}{0.11} \\

\cline{2-8}
\multirow{6}{*}{BiModal-Seq2Seq} 
& \multicolumn{1}{l}{T $\rightarrow$ V} & \multicolumn{1}{l}{{0.67}} & \multicolumn{1}{l}{\textbf{0.67}} & \multicolumn{1}{l}{\textbf{0.67}} & \multicolumn{1}{l}{{0.26}} & \multicolumn{1}{l}{{0.22}} & \multicolumn{1}{l}{\textbf{0.22}} \\
& \multicolumn{1}{l}{T $\rightarrow$ A} & \multicolumn{1}{l}{0.66} & \multicolumn{1}{l}{0.64} & \multicolumn{1}{l}{0.65} & \multicolumn{1}{l}{\textbf{0.28}} & \multicolumn{1}{l}{0.24} & \multicolumn{1}{l}{0.18} \\
& \multicolumn{1}{l}{A $\rightarrow$ T} & \multicolumn{1}{l}{0.55} & \multicolumn{1}{l}{0.60} & \multicolumn{1}{l}{0.56} & \multicolumn{1}{l}{0.17} & \multicolumn{1}{l}{\textbf{0.34}} & \multicolumn{1}{l}{0.11} \\
& \multicolumn{1}{l}{A $\rightarrow$ V} & \multicolumn{1}{l}{0.55} & \multicolumn{1}{l}{0.55} & \multicolumn{1}{l}{0.54} & \multicolumn{1}{l}{0.16} & \multicolumn{1}{l}{0.18} & \multicolumn{1}{l}{0.16} \\
& \multicolumn{1}{l}{V $\rightarrow$ T} & \multicolumn{1}{l}{0.58} & \multicolumn{1}{l}{0.58} & \multicolumn{1}{l}{0.58} & \multicolumn{1}{l}{0.05} & \multicolumn{1}{l}{0.16} & \multicolumn{1}{l}{0.08} \\
& \multicolumn{1}{l}{V $\rightarrow$ A} & \multicolumn{1}{l}{0.58} & \multicolumn{1}{l}{0.62} & \multicolumn{1}{l}{0.58} & \multicolumn{1}{l}{0.12} & \multicolumn{1}{l}{0.17} & \multicolumn{1}{l}{0.01} \\

\bottomrule
\end{tabular}
\end{center}
\caption{Bimodal results with 3 metrics: Precision, Recall and F-Score (F1)
}\label{tbl:bimodal_results}
}
\end{table*}

\begin{table*}[ht]
\small{
\begin{center}
\begin{tabular}{*8c}
\toprule
\multirow{2}{*}{Method} & \multirow{2}{*}{Feature} & \multicolumn{3}{c}{BINARY ($-1$, +1)} & \multicolumn{3}{c}{7-CLASS ($-3$, ..., +3)} \\
{} & {} & \multicolumn{1}{l}{Prec} & \multicolumn{1}{l}{Recall} & \multicolumn{1}{l}{F1} & \multicolumn{1}{l}{Prec} & \multicolumn{1}{l}{Recall} & \multicolumn{1}{l}{F1} \\

\midrule
\multirow{1}{*}{TriModal-Baseline} 
& \multicolumn{1}{l}{\texttt{concat}(T + V + A) } & \multicolumn{1}{l}{\textbf{0.75}} & \multicolumn{1}{l}{\textbf{0.75}} & \multicolumn{1}{l}{\textbf{0.75}} & \multicolumn{1}{l}{{0.24}} & \multicolumn{1}{l}{\textbf{0.27}} & \multicolumn{1}{l}{\textbf{0.24}} \\

\cline{2-8}

\multirow{6}{*}{TriModal-Seq2Seq} 
& \multicolumn{1}{l}{\texttt{embed}(T, V) $\rightarrow$ A} & \multicolumn{1}{l}{{0.56}} & \multicolumn{1}{l}{{0.60}} & \multicolumn{1}{l}{{0.57}} & \multicolumn{1}{l}{{0.10}} & \multicolumn{1}{l}{{0.16}} & \multicolumn{1}{l}{{0.09}} \\

& \multicolumn{1}{l}{\texttt{embed}(T, A) $\rightarrow$ V} & \multicolumn{1}{l}{{0.60}} & \multicolumn{1}{l}{{0.55}} & \multicolumn{1}{l}{{0.56}} & \multicolumn{1}{l}{{0.26}} & \multicolumn{1}{l}{{0.15}} & \multicolumn{1}{l}{{0.07}} \\

& \multicolumn{1}{l}{\texttt{embed}(A, V) $\rightarrow$ T} & \multicolumn{1}{l}{{0.66}} & \multicolumn{1}{l}{{0.53}} & \multicolumn{1}{l}{{0.44}} & \multicolumn{1}{l}{{0.16}} & \multicolumn{1}{l}{{0.04}} & \multicolumn{1}{l}{{0.09}} \\

& \multicolumn{1}{l}{\texttt{embed}(A, T) $\rightarrow$ V} & \multicolumn{1}{l}{{0.59}} & \multicolumn{1}{l}{{0.51}} & \multicolumn{1}{l}{{0.52}} & \multicolumn{1}{l}{{0.13}} & \multicolumn{1}{l}{{0.15}} & \multicolumn{1}{l}{{0.09}} \\

& \multicolumn{1}{l}{\texttt{embed}(V, T) $\rightarrow$ A} & \multicolumn{1}{l}{{0.59}} & \multicolumn{1}{l}{{0.60}} & \multicolumn{1}{l}{{0.59}} & \multicolumn{1}{l}{{0.11}} & \multicolumn{1}{l}{{0.17}} & \multicolumn{1}{l}{{0.10}} \\

& \multicolumn{1}{l}{\texttt{embed}(V, A) $\rightarrow$ T} & \multicolumn{1}{l}{{0.57}} & \multicolumn{1}{l}{{0.61}} & \multicolumn{1}{l}{{0.58}} & \multicolumn{1}{l}{{0.11}} & \multicolumn{1}{l}{{0.17}} & \multicolumn{1}{l}{{0.09}} \\

\cline{2-8}

& \multicolumn{1}{l}{\texttt{concat}(T, V) $\rightarrow$ A} & \multicolumn{1}{l}{{0.67}} & \multicolumn{1}{l}{{0.66}} & \multicolumn{1}{l}{{0.65}} & \multicolumn{1}{l}{{0.22}} & \multicolumn{1}{l}{{0.17}} & \multicolumn{1}{l}{{0.18}} \\

& \multicolumn{1}{l}{\texttt{concat}(A, T) $\rightarrow$ V} & \multicolumn{1}{l}{{0.54}} & \multicolumn{1}{l}{{0.55}} & \multicolumn{1}{l}{{0.63}} & \multicolumn{1}{l}{{0.19}} & \multicolumn{1}{l}{{0.15}} & \multicolumn{1}{l}{{0.21}} \\

& \multicolumn{1}{l}{\texttt{concat}(V, A) $\rightarrow$ T} & \multicolumn{1}{l}{{0.59}} & \multicolumn{1}{l}{{0.59}} & \multicolumn{1}{l}{{0.58}} & \multicolumn{1}{l}{{0.16}} & \multicolumn{1}{l}{{0.12}} & \multicolumn{1}{l}{{0.12}} \\

\cline{2-8}

& \multicolumn{1}{l}{T $\rightarrow$ \texttt{concat}(A, V)} & \multicolumn{1}{l}{{0.70}} & \multicolumn{1}{l}{{0.65}} & \multicolumn{1}{l}{{0.66}} & \multicolumn{1}{l}{{0.23}} & \multicolumn{1}{l}{{0.22}} & \multicolumn{1}{l}{{0.18}} \\

& \multicolumn{1}{l}{A $\rightarrow$ \texttt{concat}(T, V)} & \multicolumn{1}{l}{{0.55}} & \multicolumn{1}{l}{{0.53}} & \multicolumn{1}{l}{{0.54}} & \multicolumn{1}{l}{{0.18}} & \multicolumn{1}{l}{{0.20}} & \multicolumn{1}{l}{{0.18}} \\

\cline{2-8}

& \multicolumn{1}{l}{\texttt{concat}(T, A) $\rightarrow$ \texttt{concat}(T, V)} & \multicolumn{1}{l}{{0.62}} & \multicolumn{1}{l}{{0.60}} & \multicolumn{1}{l}{{0.61}} & \multicolumn{1}{l}{{0.23}} & \multicolumn{1}{l}{{0.24}} & \multicolumn{1}{l}{{0.22}} \\

& \multicolumn{1}{l}{\texttt{concat}(T, V) $\rightarrow$ \texttt{concat}(T, A)} & \multicolumn{1}{l}{0.68} & \multicolumn{1}{l}{{0.70}} & \multicolumn{1}{l}{{0.67}} & \multicolumn{1}{l}{\textbf{0.31}} & \multicolumn{1}{l}{{0.24}} & \multicolumn{1}{l}{{0.19}} \\

\bottomrule
\end{tabular}
\end{center}
\caption{Trimodal results with 3 metrics: Precision, Recall and F-Score (F1)
}\label{tbl:trimodal_results}
}
\end{table*}
%
\subsection{Baseline Unimodal Results}
We see that with the baseline model, as shown Table \ref{tbl:unimodal_baselines}, the text modality is by far the most discriminative when it comes to detecting emotion. This implies that users rely heavily on their word choice and language to convey meaning and emotion. While this may be true, we know that other works such as \citep{zadeh2018memory,poria2017context} have achieved higher scores by combining all these different modalities. This implies that with some careful thinking and pointed model construction, we should be able to improve upon our baseline unimodal results through the integration of additional modalities into our model.

\subsection{Baseline Multimodal Results}
The results of our different baseline multimodal approaches is shown in Table \ref{tbl:bimodal_results} for bimodal and Table \ref{tbl:trimodal_results} for trimodal. We see that of the multimodal baselines the model which combines the 3 modalities of text, speech, and video performed the best. The baseline model which combined text and audio arrived in second place followed closely by the combined text and video model. The model which combines video and audio arrived in last place by a significant margin. This corroborates our results from our unimodal baselines which implied that the text modality is the most discriminative modality in this dataset.

On the whole we can see that when all three modalities are working in concert we get the best result in a multimodal context, however, it is worth noting that we were not able to match out unimodal baseline with our multimodal models. This implies that there is still more to be drawn from our data when constructing our model and there is generally more work to be done. We believe that incorporating a stronger more robust representation of our data will be beneficial to our later attempts at classification. Though we view this to be out of scope of this work as the focus of this work is on learning informative representations. 
\subsection{Analysis of Baseline Failure Cases}
The common trend that we see among all of those baseline models is the consistent failure to identify extreme cases of either positive or negative emotions. We believe that this phenomenon is due to two possibilities. First we see that there are very few cases of highly positive ($+3$) and highly negative ($-3$) examples in the training data. As a result the models that are trained are highly biased towards not selecting $+3$ or $-3$ ratings. Secondly, our baseline models are performing categorical classification as opposed to regression or ordinal classification. We plan to solve by training the model to perform this type of prediction as a regression task as opposed to a categorical classification task.
\subsection{Bimodal Seq2Seq Results}
Our bimodal models require the exploration of two modalities, one for the encoding step and another for the decoding step. We explored several different different encoder/decoder frameworks for these models. The first model that we explored were representations generated from encoding exactly one modality and then decoding exactly one different modality. The results of this approach are included below in Table \ref{tbl:bimodal_results}. Here we can see that the Seq2Seq Modality Translation Model outperforms the baseline method in terms of F1 consistently and outperforms in terms of precision and recall in several cases, but not all. 
\subsection{Trimodal Seq2Seq Results}
%
We try all variations mentioned in Section \ref{subsec:variations} and 
the full breakdown of these results can be found in the Table \ref{tbl:trimodal_results}. 
According to that, while the Hierarchical Seq2Seq Modality Translation Model is a natural extension to the normal Seq2Seq Modality Translation model, it does not perform well on the CMU-MOSI dataset.
Otherwise, using the normal non-hierarchical model with concatenation variations does improve the performance, and particularly beats the baseline (for only F1 score) on the model which translates from \texttt{concat}(T,V) to \texttt{concat}(T+A) for the 7-class case. 
As mentioned in Section \ref{subsec:variations}, we favor the text (T) modality and make it repeated in this setting because it typically contributes more significantly to sentiment prediction. Indeed, we have tried to repeat video or audio modality but the result shrinks dramatically. 

One possible reason for this behavior is the scarcity of training data. Given that at every phase of Seq2Seq translation, we only have 1289 train samples, 230 validation and 269 test samples, Seq2Seq, which typically requires more data for training a good model, does not work efficiently. This affects even more in the hierarchical Seq2Seq cases where we train two phases of Seq2Seq. We project the performance will improve if we work on other dataset which is bigger, or if we pretrain our model on other dataset first before applying it to MOSI.  

%
\section{Discussion}

The language modality is the most discriminative as well as the most important towards learning multimodal representations. While we outperform the baseline multimodal approach we were unable to outperform the baseline unimodal text approach. Clearly from these results we know that that the text modality is the most discriminative of all of these modalities. However, it appears that these models which we have described are not able to truly separate the importance of the text modality. The fact that we are merging these modalities into a shared representation space is likely decreasing the resolution of the text domain and thus decreasing the modeling power of the domain. This is why we believe that the top performing multimodal model is one that incorporates the text domain so much (see Tables \ref{tbl:bimodal_results} and \ref{tbl:trimodal_results}).

It is worth noting that some of the learned representations were quite poor when it came to their use in prediction. For example, representations that were learned using only audio and video generally performed poorly. This is to be expected given the already known information that these modalities are not as discriminative as the language modality. At the same time, some of the worse performing representations were learned in the methodology of learning a representation based on an existing embedding. We believe this to be due to the representation losing the resolution of the original two domains from which the original source embedding was learned and instead being focused on learning the best representation to predict the final modality.

\section{Future Directions}
This research opens up a promising direction in joint unsupervised learning of multimodal representations and supervised learning of multimodal temporal data. We propose the following extensions that could improve performance: 

Firstly, using an Variational Autoencoder (VAE) \citep{vae_kingma} in conjunction with LSTM Encoder/Decoder model (as in the case of VAE Seq2Seq model) would be an interesting avenue to explore. This is because VAEs have been shown to learn better representations as compared to vanilla autoencoders \citep{vae_kingma,NIPS2016_6528}.

Secondly, since our method for multimodal representation learning is unsupervised, we could take advantage of larger external datasets to pre-train the multimodal representations before fine-tuning further with CMU-MOSI. We believe this will boost performance because we have limited data in CMU-MOSI for training (CMU-MOSI has 2199 training segments). Some datasets that come to mind include the Persuasion Opinion Multimodal (POM) dataset \cite{Park:2014:CAP:2663204.2663260} with 1000 total videos (longer than segments) and the IEMOCAP dataset with 10000 total segment. Since these datasets also consist of monologue speaker videos, we expect the learnt multimodal representations to generalize.

Thirdly, our method does not train our combined model end to end: the representations that we use to generated during on training run and the sentiment classification model are trained separately. Exploring an end-to-end version of this model end to end could possibly result in better performance where we could additionally fine tune the learned multimodal representation for sentiment analysis. 

\section{Conclusion}
To conclude, this paper investigate the problem of multimodal representation learning to leverage the abundance of unlabeled multimedia data available on the internet. We presente two methods for unsupervised learning of joint multimodal representations using multimodal Seq2Seq models: the \textit{Seq2Seq Modality Translation Model} and the \textit{Hierarchical Seq2Seq Modality Translation Model}. We found that these intermediate multimodal representations can then be used for multimodal downstream tasks. Our experiments indicate that the multimodal representations learned from our Seq2Seq modality translation method are highly informative and achieves improved performance on multimodal sentiment analysis. 

\section{Acknowledgements}
The authors are thankful to the many student peers who commented on and critiqued this work. Specific thanks to Louis-Phillipe Morency and Amir Zadeh for their helpful discussions and thoughtful critiques. We are grateful to our peers who helped us evaluate our methodology, in particular Stephen Tsou and Kshitij Khode. Finally, we also thank the anonymous reviewers for helpful and constructive feedback.

\bibliographystyle{acl_natbib.bst}

\bibliography{main.bib}

\begin{thebibliography}{}
\expandafter\ifx\csname natexlab\endcsname\relax\def\natexlab#1{#1}\fi

\bibitem[{Alku(1992)}]{alku1992glottal}
Paavo Alku. 1992.
\newblock Glottal wave analysis with pitch synchronous iterative adaptive
  inverse filtering.
\newblock {\em Speech communication\/} 11(2-3):109--118.

\bibitem[{Alku et~al.(2002)Alku, B{\"a}ckstr{\"o}m, and
  Vilkman}]{alku2002normalized}
Paavo Alku, Tom B{\"a}ckstr{\"o}m, and Erkki Vilkman. 2002.
\newblock Normalized amplitude quotient for parametrization of the glottal
  flow.
\newblock {\em the Journal of the Acoustical Society of America\/}
  112(2):701--710.

\bibitem[{Alku et~al.(1997)Alku, Strik, and Vilkman}]{alku1997parabolic}
Paavo Alku, Helmer Strik, and Erkki Vilkman. 1997.
\newblock Parabolic spectral parameter - a new method for quantification of the
  glottal flow.
\newblock {\em Speech Communication\/} 22(1):67--79.

\bibitem[{Bahdanau et~al.(2014)Bahdanau, Cho, and Bengio}]{bahdanau2014neural}
Dzmitry Bahdanau, Kyunghyun Cho, and Yoshua Bengio. 2014.
\newblock Neural machine translation by jointly learning to align and
  translate.
\newblock {\em arXiv preprint arXiv:1409.0473\/} .

\bibitem[{Baltrusaitis et~al.(2017)Baltrusaitis, Ahuja, and
  Morency}]{mm_survey}
Tadas Baltrusaitis, Chaitanya Ahuja, and Louis{-}Philippe Morency. 2017.
\newblock \href{http://arxiv.org/abs/1705.09406}{Multimodal machine learning:
  {A} survey and taxonomy}.
\newblock {\em CoRR\/} abs/1705.09406.
\newblock
  \href{http://arxiv.org/abs/1705.09406}{http://arxiv.org/abs/1705.09406}.

\bibitem[{Chaumartin(2007)}]{Chaumartin_2007}
Fran\c{c}ois-R{\'e}gis Chaumartin. 2007.
\newblock \href{http://dl.acm.org/citation.cfm?id=1621474.1621568}{Upar7: A
  knowledge-based system for headline sentiment tagging}.
\newblock In {\em Proceedings of the 4th International Workshop on Semantic
  Evaluations\/}. Association for Computational Linguistics, Stroudsburg, PA,
  USA, SemEval '07, pages 422--425.
\newblock
  \href{http://dl.acm.org/citation.cfm?id=1621474.1621568}{http://dl.acm.org/citation.cfm?id=1621474.1621568}.

\bibitem[{Chen et~al.(2017)Chen, Wang, Liang, Baltrusaitis, Zadeh, and
  Louis-Phillippe}]{chen2017msa}
Minghai Chen, Sen Wang, Paul~Pu Liang, Tadas Baltrusaitis, Amir Zadeh, and
  Morency Louis-Phillippe. 2017.
\newblock Multimodal sentiment analysis with word-level fusion and
  reinforcement learning.
\newblock {\em ICMI, Glassgow, United Kingdom\/} .

\bibitem[{Childers and Lee(1991)}]{childers1991vocal}
Donald~G Childers and CK~Lee. 1991.
\newblock Vocal quality factors: Analysis, synthesis, and perception.
\newblock {\em the Journal of the Acoustical Society of America\/}
  90(5):2394--2410.

\bibitem[{Chung et~al.(2015)Chung, Gulcehre, Cho, and Bengio}]{gru}
Junyoung Chung, Caglar Gulcehre, Kyunghyun Cho, and Yoshua Bengio. 2015.
\newblock Gated feedback recurrent neural networks.
\newblock In {\em International Conference on Machine Learning\/}. pages
  2067--2075.

\bibitem[{Degottex et~al.(2014)Degottex, Kane, Drugman, Raitio, and
  Scherer}]{degottex2014covarep}
Gilles Degottex, John Kane, Thomas Drugman, Tuomo Raitio, and Stefan Scherer.
  2014.
\newblock Covarep - a collaborative voice analysis repository for speech
  technologies.
\newblock In {\em Acoustics, Speech and Signal Processing (ICASSP), 2014 IEEE
  International Conference on\/}. IEEE, pages 960--964.

\bibitem[{Donahue et~al.(2016)Donahue, Kr{\"a}henb{\"u}hl, and Darrell}]{bigan}
Jeff Donahue, Philipp Kr{\"a}henb{\"u}hl, and Trevor Darrell. 2016.
\newblock Adversarial feature learning.
\newblock {\em arXiv preprint arXiv:1605.09782\/} .

\bibitem[{Drugman and Alwan(2011)}]{drugman2011joint}
Thomas Drugman and Abeer Alwan. 2011.
\newblock Joint robust voicing detection and pitch estimation based on residual
  harmonics.
\newblock In {\em Interspeech\/}. pages 1973--1976.

\bibitem[{Drugman et~al.(2012)Drugman, Thomas, Gudnason, Naylor, and
  Dutoit}]{drugman2012detection}
Thomas Drugman, Mark Thomas, Jon Gudnason, Patrick Naylor, and Thierry Dutoit.
  2012.
\newblock Detection of glottal closure instants from speech signals: A
  quantitative review.
\newblock {\em IEEE Transactions on Audio, Speech, and Language Processing\/}
  20(3):994--1006.

\bibitem[{Ekman(1992)}]{ekman1992argument}
Paul Ekman. 1992.
\newblock An argument for basic emotions.
\newblock {\em Cognition \& emotion\/} 6(3-4):169--200.

\bibitem[{Ekman et~al.(1980)Ekman, Freisen, and Ancoli}]{ekman1980facial}
Paul Ekman, Wallace~V Freisen, and Sonia Ancoli. 1980.
\newblock Facial signs of emotional experience.
\newblock {\em Journal of personality and social psychology\/} 39(6):1125.

\bibitem[{Gan et~al.(2017)Gan, Chen, Wang, Pu, Zhang, Liu, Li, and
  Carin}]{trianglegan}
Zhe Gan, Liqun Chen, Weiyao Wang, Yunchen Pu, Yizhe Zhang, Hao Liu, Chunyuan
  Li, and Lawrence Carin. 2017.
\newblock Triangle generative adversarial networks.
\newblock {\em arXiv preprint arXiv:1709.06548\/} .

\bibitem[{Goodfellow et~al.(2014)Goodfellow, Pouget-Abadie, Mirza, Xu,
  Warde-Farley, Ozair, Courville, and Bengio}]{gan}
Ian Goodfellow, Jean Pouget-Abadie, Mehdi Mirza, Bing Xu, David Warde-Farley,
  Sherjil Ozair, Aaron Courville, and Yoshua Bengio. 2014.
\newblock Generative adversarial nets.
\newblock In {\em Advances in neural information processing systems\/}. pages
  2672--2680.

\bibitem[{Hochreiter and Schmidhuber(1997)}]{lstm}
Sepp Hochreiter and J{\"u}rgen Schmidhuber. 1997.
\newblock Long short-term memory.
\newblock {\em Neural computation\/} 9(8):1735--1780.

\bibitem[{iMotions(2017)}]{emotient}
iMotions. 2017.
\newblock \href{goo.gl/1rh1JN}{Facial expression analysis}.
\newblock \href{goo.gl/1rh1JN}{goo.gl/1rh1JN}.

\bibitem[{Kane and Gobl(2013)}]{kane2013wavelet}
John Kane and Christer Gobl. 2013.
\newblock Wavelet maxima dispersion for breathy to tense voice discrimination.
\newblock {\em IEEE Transactions on Audio, Speech, and Language Processing\/}
  21(6):1170--1179.

\bibitem[{Kaushik et~al.(2013)Kaushik, Sangwan, and
  Hansen}]{kaushik2013sentiment}
Lakshmish Kaushik, Abhijeet Sangwan, and John~HL Hansen. 2013.
\newblock Sentiment extraction from natural audio streams.
\newblock In {\em Acoustics, speech and signal processing (icassp), 2013 ieee
  international conference on\/}. IEEE, pages 8485--8489.

\bibitem[{Kingma et~al.(2014)Kingma, Mohamed, Rezende, and
  Welling}]{conditionalvae}
Diederik~P Kingma, Shakir Mohamed, Danilo~Jimenez Rezende, and Max Welling.
  2014.
\newblock Semi-supervised learning with deep generative models.
\newblock In {\em Advances in Neural Information Processing Systems\/}. pages
  3581--3589.

\bibitem[{Kingma and Welling(2013)}]{vae_kingma}
Diederik~P Kingma and Max Welling. 2013.
\newblock Auto-encoding variational bayes.
\newblock {\em arXiv preprint arXiv:1312.6114\/} .

\bibitem[{Kiros et~al.(2014)Kiros, Salakhutdinov, and
  Zemel}]{kiros2014unifying}
Ryan Kiros, Ruslan Salakhutdinov, and Richard~S Zemel. 2014.
\newblock Unifying visual-semantic embeddings with multimodal neural language
  models.
\newblock {\em arXiv preprint arXiv:1411.2539\/} .

\bibitem[{{Klein} et~al.(){Klein}, {Kim}, {Deng}, {Senellart}, and
  {Rush}}]{2017opennmt}
G.~{Klein}, Y.~{Kim}, Y.~{Deng}, J.~{Senellart}, and A.~M. {Rush}. ????
\newblock {OpenNMT: Open-Source Toolkit for Neural Machine Translation}.
\newblock {\em ArXiv e-prints\/} .

\bibitem[{Lazaridou et~al.(2015)Lazaridou, Pham, and
  Baroni}]{lazaridou2015combining}
Angeliki Lazaridou, Nghia~The Pham, and Marco Baroni. 2015.
\newblock Combining language and vision with a multimodal skip-gram model.
\newblock {\em arXiv preprint arXiv:1501.02598\/} .

\bibitem[{Li et~al.(2017)Li, Xu, Zhu, and Zhang}]{triplegan}
Chongxuan Li, Kun Xu, Jun Zhu, and Bo~Zhang. 2017.
\newblock Triple generative adversarial nets.
\newblock {\em arXiv preprint arXiv:1703.02291\/} .

\bibitem[{Luong et~al.(2015)Luong, Pham, and Manning}]{luong2015effective}
Minh-Thang Luong, Hieu Pham, and Christopher~D Manning. 2015.
\newblock Effective approaches to attention-based neural machine translation.
\newblock {\em arXiv preprint arXiv:1508.04025\/} .

\bibitem[{Mirza and Osindero(2014)}]{conditionalgan}
Mehdi Mirza and Simon Osindero. 2014.
\newblock Conditional generative adversarial nets.
\newblock {\em arXiv preprint arXiv:1411.1784\/} .

\bibitem[{Mishne et~al.(2005)}]{mishne2005experiments}
Gilad Mishne et~al. 2005.
\newblock Experiments with mood classification in blog posts.
\newblock In {\em Proceedings of ACM SIGIR 2005 workshop on stylistic analysis
  of text for information access\/}. volume~19, pages 321--327.

\bibitem[{Morency et~al.(2011)Morency, Mihalcea, and
  Doshi}]{morency2011towards}
Louis-Philippe Morency, Rada Mihalcea, and Payal Doshi. 2011.
\newblock Towards multimodal sentiment analysis: Harvesting opinions from the
  web.
\newblock In {\em Proceedings of the 13th international conference on
  multimodal interfaces\/}. ACM, pages 169--176.

\bibitem[{Neubig(2017)}]{neubig2017neural}
Graham Neubig. 2017.
\newblock Neural machine translation and sequence-to-sequence models: A
  tutorial.
\newblock {\em arXiv preprint arXiv:1703.01619\/} .

\bibitem[{Ngiam et~al.(2011)Ngiam, Khosla, Kim, Nam, Lee, and
  Ng}]{ngiam2011multimodal}
Jiquan Ngiam, Aditya Khosla, Mingyu Kim, Juhan Nam, Honglak Lee, and Andrew~Y
  Ng. 2011.
\newblock Multimodal deep learning.
\newblock In {\em Proceedings of the 28th international conference on machine
  learning (ICML-11)\/}. pages 689--696.

\bibitem[{Pandey and Dukkipati(2017)}]{variationalmultimodal}
Gaurav Pandey and Ambedkar Dukkipati. 2017.
\newblock Variational methods for conditional multimodal deep learning.
\newblock In {\em Neural Networks (IJCNN), 2017 International Joint Conference
  on\/}. IEEE, pages 308--315.

\bibitem[{Park et~al.(2014)Park, Shim, Chatterjee, Sagae, and
  Morency}]{Park:2014:CAP:2663204.2663260}
Sunghyun Park, Han~Suk Shim, Moitreya Chatterjee, Kenji Sagae, and
  Louis-Philippe Morency. 2014.
\newblock \href{https://doi.org/10.1145/2663204.2663260}{Computational analysis
  of persuasiveness in social multimedia: A novel dataset and multimodal
  prediction approach}.
\newblock In {\em Proceedings of the 16th International Conference on
  Multimodal Interaction\/}. ACM, New York, NY, USA, ICMI '14, pages 50--57.
\newblock
  \href{https://doi.org/10.1145/2663204.2663260}{https://doi.org/10.1145/2663204.2663260}.

\bibitem[{Pennington et~al.(2014)Pennington, Socher, and
  Manning}]{pennington2014glove}
Jeffrey Pennington, Richard Socher, and Christopher~D Manning. 2014.
\newblock Glove: Global vectors for word representation.
\newblock In {\em EMNLP\/}. volume~14, pages 1532--1543.

\bibitem[{Poria et~al.(2017)Poria, Cambria, Hazarika, Majumder, Zadeh, and
  Morency}]{poria2017context}
Soujanya Poria, Erik Cambria, Devamanyu Hazarika, Navonil Majumder, Amir Zadeh,
  and Louis-Philippe Morency. 2017.
\newblock Context-dependent sentiment analysis in user-generated videos.
\newblock In {\em Proceedings of the 55th Annual Meeting of the Association for
  Computational Linguistics (Volume 1: Long Papers)\/}. volume~1, pages
  873--883.

\bibitem[{Pu et~al.(2016)Pu, Gan, Henao, Yuan, Li, Stevens, and
  Carin}]{NIPS2016_6528}
Yunchen Pu, Zhe Gan, Ricardo Henao, Xin Yuan, Chunyuan Li, Andrew Stevens, and
  Lawrence Carin. 2016.
\newblock Variational autoencoder for deep learning of images, labels and
  captions.
\newblock In D.~D. Lee, M.~Sugiyama, U.~V. Luxburg, I.~Guyon, and R.~Garnett,
  editors, {\em Advances in Neural Information Processing Systems 29\/}, Curran
  Associates, Inc., pages 2352--2360.

\bibitem[{Rajagopalan et~al.(2016)Rajagopalan, Morency, Baltrusaitis, and
  Goecke}]{Rajagopalan2016}
Shyam~Sundar Rajagopalan, Louis-Philippe Morency, Tadas Baltrusaitis, and
  Roland Goecke. 2016.
\newblock {\em Extending Long Short-Term Memory for Multi-View Structured
  Learning\/}, Springer International Publishing, Cham, pages 338--353.

\bibitem[{Rosas et~al.(2013)Rosas, Mihalcea, and Morency}]{rosas2013multimodal}
Ver{\'o}nica~P{\'e}rez Rosas, Rada Mihalcea, and Louis-Philippe Morency. 2013.
\newblock Multimodal sentiment analysis of spanish online videos.
\newblock {\em IEEE Intelligent Systems\/} 28(3):38--45.

\bibitem[{Sriram et~al.(2017)Sriram, Jun, Satheesh, and Coates}]{coldfusion}
Anuroop Sriram, Heewoo Jun, Sanjeev Satheesh, and Adam Coates. 2017.
\newblock Cold fusion: Training seq2seq models together with language models.
\newblock {\em arXiv preprint arXiv:1708.06426\/} .

\bibitem[{Sutskever et~al.(2014)Sutskever, Vinyals, and Le}]{Seq2Seq}
Ilya Sutskever, Oriol Vinyals, and Quoc~V Le. 2014.
\newblock Sequence to sequence learning with neural networks.
\newblock In {\em Advances in neural information processing systems\/}. pages
  3104--3112.

\bibitem[{W{\"o}llmer et~al.(2013)W{\"o}llmer, Weninger, Knaup, Schuller, Sun,
  Sagae, and Morency}]{wollmer2013youtube}
Martin W{\"o}llmer, Felix Weninger, Tobias Knaup, Bj{\"o}rn Schuller, Congkai
  Sun, Kenji Sagae, and Louis-Philippe Morency. 2013.
\newblock Youtube movie reviews: Sentiment analysis in an audio-visual context.
\newblock {\em IEEE Intelligent Systems\/} 28(3):46--53.

\bibitem[{Xu et~al.(2015)Xu, Ba, Kiros, Cho, Courville, Salakhudinov, Zemel,
  and Bengio}]{xu2015show}
Kelvin Xu, Jimmy Ba, Ryan Kiros, Kyunghyun Cho, Aaron Courville, Ruslan
  Salakhudinov, Rich Zemel, and Yoshua Bengio. 2015.
\newblock Show, attend and tell: Neural image caption generation with visual
  attention.
\newblock In {\em International Conference on Machine Learning\/}. pages
  2048--2057.

\bibitem[{Yuan and Liberman(2008)}]{P2FA}
Jiahong Yuan and Mark Liberman. 2008.
\newblock Speaker identification on the scotus corpus.
\newblock {\em Journal of the Acoustical Society of America\/} 123(5):3878.

\bibitem[{Zadeh et~al.(2017)Zadeh, Chen, Poria, Cambria, and
  Morency}]{tensoremnlp17}
Amir Zadeh, Minghai Chen, Soujanya Poria, Erik Cambria, and Louis-Philippe
  Morency. 2017.
\newblock Tensor fusion network for multimodal sentiment analysis.
\newblock In {\em Proceedings of the 2017 Conference on Empirical Methods in
  Natural Language Processing\/}. pages 1114--1125.

\bibitem[{Zadeh et~al.(2018{\natexlab{a}})Zadeh, Liang, Mazumder, Poria,
  Cambria, and Morency}]{zadeh2018memory}
Amir Zadeh, Paul~Pu Liang, Navonil Mazumder, Soujanya Poria, Erik Cambria, and
  Louis-Philippe Morency. 2018{\natexlab{a}}.
\newblock Memory fusion network for multi-view sequential learning.
\newblock {\em arXiv preprint arXiv:1802.00927\/} .

\bibitem[{Zadeh et~al.(2018{\natexlab{b}})Zadeh, Liang, Poria, Vij, Cambria,
  and Morency}]{zadeh2018multi}
Amir Zadeh, Paul~Pu Liang, Soujanya Poria, Prateek Vij, Erik Cambria, and
  Louis-Philippe Morency. 2018{\natexlab{b}}.
\newblock Multi-attention recurrent network for human communication
  comprehension.
\newblock {\em arXiv preprint arXiv:1802.00923\/} .

\bibitem[{Zadeh et~al.(2018{\natexlab{c}})Zadeh, Liang, Vanbriesen, Poria,
  Cambria, Chen, and Morency}]{cmumoseiacl2018}
Amir Zadeh, Paul~Pu Liang, Jon Vanbriesen, Soujanya Poria, Erik Cambria,
  Minghai Chen, and Louis-Philippe Morency. 2018{\natexlab{c}}.
\newblock Multimodal language analysis in the wild: Cmu-mosei dataset and
  interpretable dynamic fusion graph.
\newblock In {\em Association for Computational Linguistics (ACL)\/}.

\bibitem[{Zadeh et~al.(2016)Zadeh, Zellers, Pincus, and
  Morency}]{zadeh2016multimodal}
Amir Zadeh, Rowan Zellers, Eli Pincus, and Louis-Philippe Morency. 2016.
\newblock Multimodal sentiment intensity analysis in videos: Facial gestures
  and verbal messages.
\newblock {\em IEEE Intelligent Systems\/} 31(6):82--88.

\end{thebibliography}

\end{document}